\pdfoutput=1
\documentclass[11pt]{article}

\usepackage[a4paper,margin=22mm,top=22mm,bottom=22mm]{geometry}
\usepackage[T1]{fontenc}
\usepackage[utf8]{inputenc}
\usepackage{amsmath}
\usepackage{newtxtext,newtxmath}
\usepackage{microtype}
\usepackage{graphicx}
\usepackage{caption}
\usepackage{setspace}
\usepackage{titlesec}
\usepackage{indentfirst}
\usepackage[noadjust]{cite}
\usepackage[hidelinks]{hyperref}
\hypersetup{
  pdftitle={Asking the right question is key to AI discovery in a chain of hypotheses},
  pdfauthor={Hu Cang}
}

\makeatletter
\renewcommand\@cite[2]{\textsuperscript{#1}}

\renewcommand\@biblabel[1]{#1.}
\makeatother

\titleformat{\section}{\normalfont\large\bfseries}{\thesection}{0.5em}{}
\titleformat{\subsection}{\normalfont\normalsize\bfseries}{\thesubsection}{0.5em}{}
\titlespacing*{\section}{0pt}{9.5pt}{2.5pt}
\titlespacing*{\subsection}{0pt}{7pt}{2pt}

\begin{document}

\begin{center}
{\LARGE\bfseries Asking the right question is key to AI discovery\\[2pt] in a chain of hypotheses\par}
\vspace{8pt}
{\normalsize Hu Cang$^{1,*}$\par}
\vspace{3pt}
{\small $^{1}$Department of Developmental and Cell Biology, University of California, Irvine, CA 92697, USA.\par}
{\small $^{*}$e-mail: cangh@uci.edu\par}
\end{center}
\vspace{4pt}

\section*{Abstract}

Artificial intelligence (AI) scientists pursue discovery through a chain of hypotheses, fitting candidate models to existing observations and often choosing experiments where those models disagree most. Yet when early data miss a narrow feature, model selection favors simple curves that agree across the unmeasured gap, so disagreement-guided search has no reason to look where the new phenomenon lies. When we tested two frontier AI scientists, Claude Sonnet 5.5 and GPT-6.1-Sol, across 32 closed-loop trials with a narrow peak omitted from the starting data, both sampled inside the peak region less often than random search. Surprisingly, the agent that followed model disagreement more closely missed the peak in all 32 trials, whereas discovery succeeded when experiments probed the unmeasured gap early and followed up across the anomaly. Asking the right question, rather than testing only where current models disagree, is therefore key to AI discovery in a chain of hypotheses.

\section*{Introduction}

Scientific inquiry advances through a chain of hypotheses: investigators propose explanations for what they have observed, run experiments to test them, and revise their models when measurements depart from expectation\cite{chamberlin1890method,platt1964strong,king2009automation,wang2023scientific}. Autonomous language-model agents can now run this loop end to end, proposing hypotheses and choosing which experiments to perform next\cite{boiko2023coscientist,szymanski2023alab,lu2024aiscientist,gottweis2025coscientist,kabra2026autoscilab}. Most efforts to improve AI scientists have focused on building stronger equation and code generators\cite{schmidt2009distilling,cranmer2023interpretable,romera2024funsearch,novikov2025alphaevolve} on the assumption that the best next experiment is simply one that distinguishes among current candidate models. In classical experimental design and active learning, a standard strategy is measuring where the leading models disagree most\cite{lindley1956measure,box1967discrimination,atkinson1975design,seung1992query,chaloner1995bayesian,settles2009active,rainforth2024modern}.

Yet measuring where models disagree works only when those models already differ where new phenomena lie\cite{sugiyama2006active,sloman2022characterizing,tang2026ridea}. Early in an investigation, sparse observations can miss a localized feature altogether, and standard model selection penalizes extra parameters until the data show a clear departure\cite{schwarz1978estimating}. Simple models fitted to the observed background then agree across the unmeasured gap and differ only over minor details elsewhere. Following their disagreement steers experiments away from that gap rather than into it, much as confirmation bias keeps human researchers testing what their working theories already anticipate\cite{wason1960failure,klayman1987confirmation,kaplan1964conduct}.

To test how experimental choices shape autonomous discovery, we evaluated two frontier AI scientists, Claude Sonnet 5.5 and GPT-6.1-Sol, in closed-loop equation-discovery trials where a narrow peak lay inside an initially unmeasured region. Both agents probed that region less often than random search, and the agent that followed equation disagreement more closely missed the peak in every trial as it spent its budget separating nearly identical background curves. By contrast, discovery succeeded when experiments entered the unmeasured gap early, detected a clear anomaly, and completed a multi-step chain of hypotheses across the peak. Asking the right question is therefore critical to AI discovery.

\section*{Results}

\subsection*{Testing discovery when initial data miss a narrow feature}

\begin{figure}[!t]
  \centering
  \includegraphics[width=\textwidth]{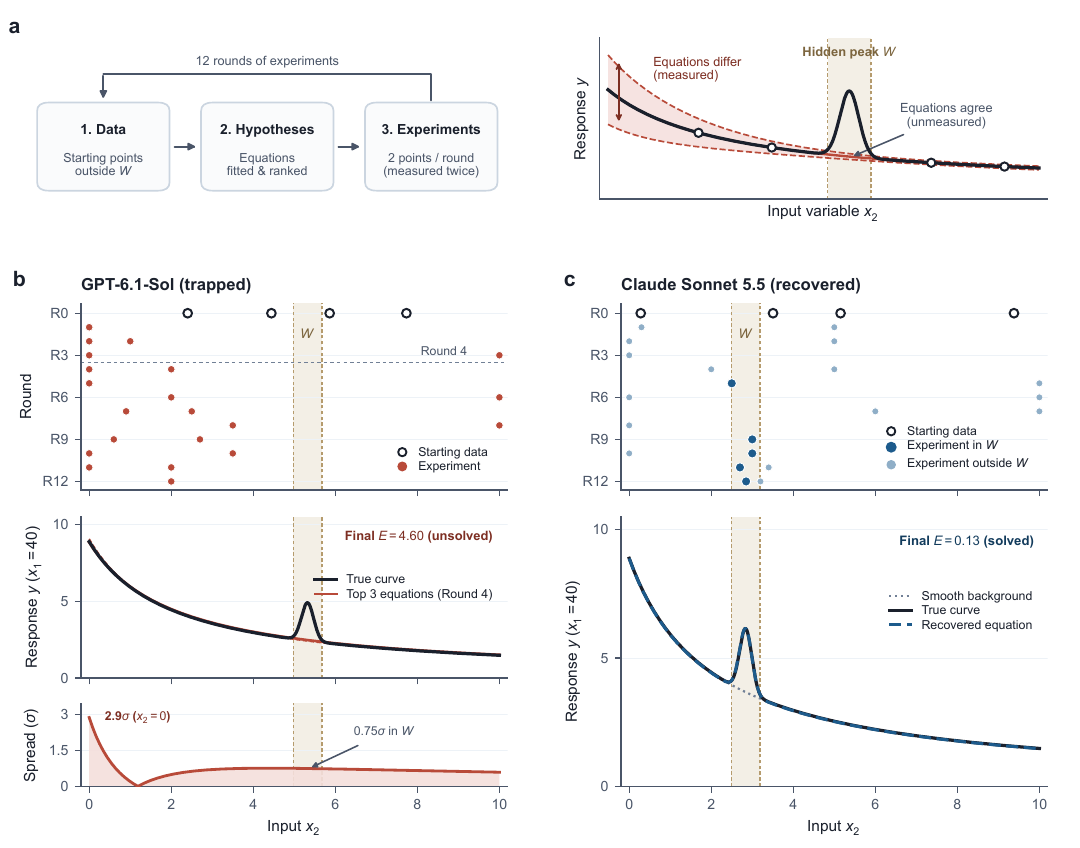}
  \caption{\textbf{How unmeasured gaps trap a closed-loop chain of hypotheses.}
  \textbf{a}, Study design and blind-spot mechanism. Left: over 12 rounds, the agent proposes candidate equations for existing data and chooses two new conditions per round (each measured twice). Right: when starting data (open circles) miss a narrow peak region $W$ (shaded band), smooth curves fitted to the background differ near $x_2 = 0$ but agree across the unmeasured gap, steering disagreement-based experiments away from $W$.
  \textbf{b}, Example of a trapped trial (GPT-6.1-Sol). Top: chosen $x_2$ values across Rounds 1--12 never enter $W$. Middle and bottom: by Round 4, the top three equations agree within $0.75\sigma$ across $W$ and differ mainly at $x_2 = 0$ ($2.9\sigma$), leaving the agent on a smooth curve without the peak (final error $E = 4.60$).
  \textbf{c}, Example of a successful recovery (Claude Sonnet 5.5). Top: after a weak edge point in Round 5 ($1.6\sigma$), a strong measurement inside $W$ in Round 9 ($x_2 = 3.0$, $15.7\sigma$) reveals the jump, prompting a peak term in Round 10 and follow-up measurements across $W$. Bottom: the final equation matches the true curve ($E = 0.13$).}
  \label{fig:one}
\end{figure}

In our closed-loop task, Claude Sonnet 5.5 and GPT-6.1-Sol sought to discover an unknown equation $y = f(x_1, x_2, x_3) + \varepsilon$ over 12 rounds of experiments (Fig.~\ref{fig:one}a and Methods). In this system, $y$ rises and levels off with $x_1$, decays smoothly with $x_2$, and does not depend on $x_3$. Hidden inside that smooth decay is a narrow Gaussian peak region $W$ covering $7\%$ of the $x_2$ range. At $x_1 = 40$, $y$ rises up to $31\sigma$ above the smooth curve inside $W$ (in units of measurement noise $\sigma$) while adding less than $2\sigma$ anywhere outside $W$. Across 32 paired trials with different peak locations along $x_2$, both agents received the same four random starting conditions outside $W$, so the initial data showed only the smooth background.

In each round, the agent saw all measurements collected so far alongside its best-fitting equations from earlier rounds, ranked by the Bayesian information criterion (BIC)\cite{schwarz1978estimating}, which rewards fit accuracy while penalizing extra parameters. The agent then proposed up to three new equations and chose two input conditions to measure next. After Round 12, the agent could submit up to three final equations without further measurements, and a trial counted as solved if the top-ranked equation predicted held-out test points within measurement noise (root-mean-square error $E = \mathrm{RMSE}/\sigma \le 1.05$). Two example trials illustrate how experimental choices decide the outcome (Fig.~\ref{fig:one}b,c). In one trial, once GPT-6.1-Sol fits smooth decay curves to the starting data, its top three equations agree closely across the unmeasured gap around $W$ and differ mainly near $x_2 = 0$. Because the agent tests where its leading equations disagree, it keeps measuring near $x_2 = 0$, never enters $W$, and ends on a smooth curve without the peak (Fig.~\ref{fig:one}b). In another trial, after a weak point near the edge of $W$ in Round 5 leaves the smooth curve intact, a strong measurement inside $W$ in Round 9 reveals a large jump that smooth curves cannot explain. That departure prompts Sonnet 5.5 to add a peak term in Round 10 and place follow-up measurements across $W$ until the true curve is recovered (Fig.~\ref{fig:one}c).

\subsection*{Agreement among incomplete equations steers search away from the peak}

\begin{figure}[!t]
  \centering
  \includegraphics[width=\textwidth]{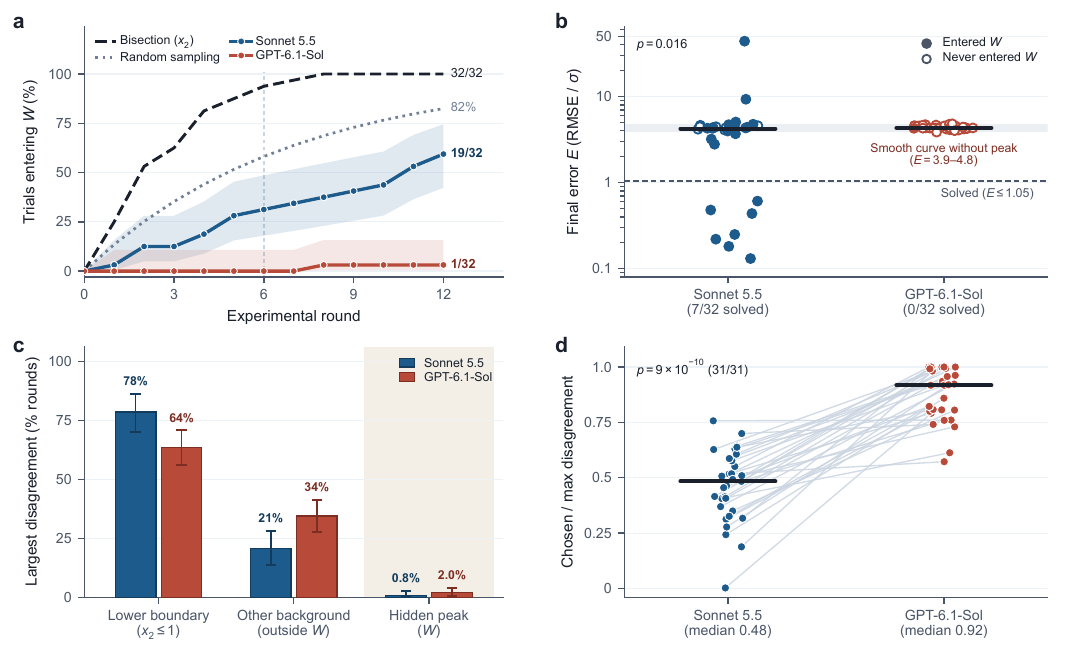}
  \caption{\textbf{Both AI scientists measure inside the hidden peak less often than simple search baselines.}
  \textbf{a}, Percentage of the 32 paired trials entering $W$ by each round, compared with two search strategies that fit no equations: interval bisection along $x_2$ (repeatedly measuring the midpoint of the widest unmeasured interval) and random sampling. Shaded bands show $95\%$ confidence intervals; the vertical dashed line marks Round 6 ($p = 0.002$ for Sonnet 5.5 and $p = 8 \times 10^{-13}$ for GPT-6.1-Sol below random sampling).
  \textbf{b}, Final prediction error $E = \mathrm{RMSE}/\sigma$ on held-out test points across all 32 paired trials (horizontal bars, medians). Trials below the dashed line ($E \le 1.05$) are solved ($7/32$ for Sonnet 5.5 versus $0/32$ for GPT-6.1-Sol; $p = 0.016$). Smooth curves without the peak fall inside the gray band ($E = 3.9$--$4.8$).
  \textbf{c}, Where each agent's top three equations disagreed most during rounds before entering $W$ ($251$ rounds for Sonnet 5.5; $348$ for GPT-6.1-Sol). Disagreement concentrated near $x_2 \le 1$ and rarely fell inside $W$ ($95\%$ bootstrap confidence intervals).
  \textbf{d}, Ratio of equation disagreement at the chosen experiment to the maximum disagreement anywhere in the space (per-trial median across $n = 31$ shared trials; $1$ = testing at maximum disagreement). Gray lines connect the same trial across agents, and horizontal bars mark medians ($p = 9 \times 10^{-10}$; one trial excluded because Sonnet 5.5 entered $W$ in Round 1).}
  \label{fig:two}
\end{figure}

Across all 32 paired trials, we first asked how often each agent's experiments even entered the peak region $W$. We compared both agents against two simple search baselines that fit no equations at all: random sampling across $x_2$ and interval bisection along $x_2$ (Fig.~\ref{fig:two}a). Interval bisection repeatedly measures the midpoint of the widest unmeasured interval in $x_2$, using prior knowledge of which variable carries the peak\cite{johnson1990minimax}. Random sampling would enter $W$ in $82\%$ of trials by Round 12, and interval bisection entered all $32/32$. Both AI scientists entered $W$ less often than random sampling, falling significantly below random already by Round 6 ($p = 0.002$ for Sonnet 5.5 and $p = 8 \times 10^{-13}$ for GPT-6.1-Sol). Over all 12 rounds, Sonnet 5.5 entered $W$ in $19/32$ trials ($59\%$) and GPT-6.1-Sol in just $1/32$ ($3\%$). Because neither agent can fit a feature it never observes, missing $W$ prevented recovery (Fig.~\ref{fig:two}b). Sonnet 5.5 solved $7/32$ trials ($22\%$) versus $0/32$ for GPT-6.1-Sol ($p = 0.016$), whose runs all ended on smooth curves without a peak.

Inspecting the agents' candidate equations before any measurement entered $W$ explains why equation-guided search performed worse than random sampling (Fig.~\ref{fig:two}c,d). Without a data point inside $W$, adding peak parameters worsens BIC rather than improving it. Neither agent's top-ranked equation predicted a peak in more than $1\%$ of rounds before entering $W$, and both agents rarely proposed peak terms until a measurement landed in $W$ (Methods). Meanwhile, the smooth decay curves that led the rankings were steepest near $x_2 = 0$ and nearly flat across the interior. Small parameter differences among the top three equations therefore produced their largest prediction spread near $x_2 \le 1$ ($78\%$ of rounds for Sonnet 5.5 and $64\%$ for GPT-6.1-Sol) and almost never inside $W$ ($0.8\%$ and $2.0\%$; Fig.~\ref{fig:two}c).

This also explains why GPT-6.1-Sol missed the peak in almost every trial (Fig.~\ref{fig:two}d). For each round, we divided the disagreement among the top three equations at the chosen point by their maximum disagreement anywhere in the space, giving a ratio from $0$ to $1$ ($1$ means testing right at maximum disagreement). GPT-6.1-Sol scored higher than Sonnet 5.5 in all $31$ shared trials (median $0.92$ versus $0.48$; $p = 9 \times 10^{-10}$). Yet across those $348$ rounds before entering $W$, GPT-6.1-Sol's top three equations differed at most by a median of $0.68\sigma$ ($1.25\sigma$ across $251$ rounds for Sonnet 5.5), less than single-measurement noise. By consistently testing where those equations disagreed, GPT-6.1-Sol spent its rounds trying to separate nearly identical background curves and left interior gaps untested.

\subsection*{Testing inside $W$ starts a multi-step chain of hypotheses}

\begin{figure}[!t]
  \centering
  \includegraphics[width=\textwidth]{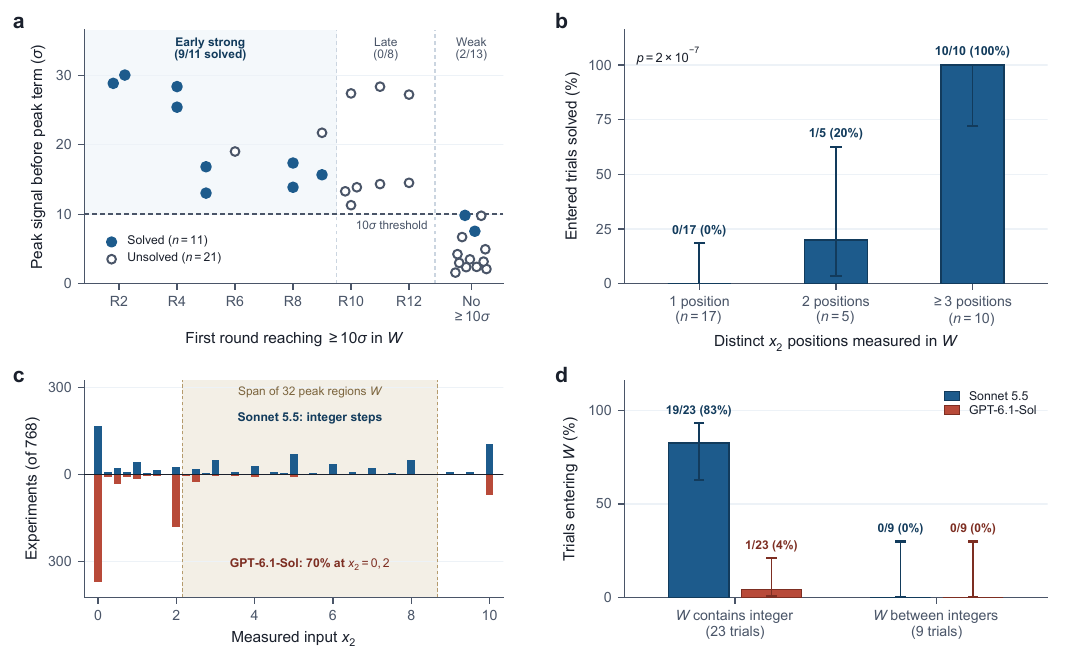}
  \caption{\textbf{Recovering the hidden peak requires a multi-step chain of hypotheses after entering $W$.}
  \textbf{a}, Anomaly strength and timing across the $32$ Sonnet 5.5 trials that entered $W$ (out of $64$ total runs). Each point shows the strongest peak signal (in units of $\sigma$) measured in $W$ before a peak term entered the top-ranked equation (or through Round 12 if none did), plotted against the first round reaching $\ge 10\sigma$.
  \textbf{b}, Percentage of the $32$ entered Sonnet 5.5 trials solved by the number of distinct $x_2$ positions measured inside $W$ ($1$, $2$, or $\ge 3$ positions; $p = 2 \times 10^{-7}$ for $\ge 3$ versus $1$--$2$ positions; error bars, $95\%$ confidence intervals).
  \textbf{c}, Distribution of all $768$ measured $x_2$ values across the $32$ paired trials for Sonnet 5.5 (top) and GPT-6.1-Sol (bottom) in $0.25$-unit bins; the shaded band spans all $32$ peak regions $W$.
  \textbf{d}, Percentage of the $32$ paired trials entering $W$ by Round 12 when $W$ contained an integer ($23$ trials) versus lay between consecutive integers ($9$ trials; error bars, $95\%$ confidence intervals).}
  \label{fig:three}
\end{figure}

Across both agents, asking what happens inside $W$ was the first key to discovery: neither agent ever recovered the peak without measuring there. Without a data point in the gap, the chain of hypotheses had no evidence to move beyond the smooth curve. Because GPT-6.1-Sol entered $W$ only once, we analyzed post-entry discovery across $64$ Sonnet 5.5 trials run under the same protocol ($32$ of which entered $W$, with $0$ of $32$ solved without entering $W$; Methods). Entering $W$ enabled Sonnet 5.5 to solve $11$ of those $32$ trials ($21$ unsolved). Once experiments reached the gap, the agent could recover the full law when supported by a chain of follow-up measurements (Fig.~\ref{fig:three}a,b).

First, the measurement inside $W$ had to stand out clearly from the smooth background before BIC would favor adding extra parameters (Fig.~\ref{fig:three}a). Because the peak tapers toward the edges of $W$ and scales with $x_1$, experiments near the boundary of $W$ or at low $x_1$ produced only modest departures. When a measurement hit the peak strongly ($\ge 10\sigma$, $19$ of the $32$ entered trials), the large error overturned the smooth fit and launched a peak term in the top-ranked equation in $16$ of the $19$ trials (Methods). Among the $13$ entered trials whose initial signals stayed below $10\sigma$ (rightmost column in Fig.~\ref{fig:three}a), only $2$ solved ($7.6\sigma$ and $9.8\sigma$ across multiple points in $W$). The other $11$ never added a peak term to the top-ranked equation.

Second, once a strong hit launched a peak hypothesis, further steps in the hypothesis chain pinned down its shape (Fig.~\ref{fig:three}b). One $x_2$ point inside $W$ cannot determine a peak's height, center, and width at once. Across the $32$ entered trials, Sonnet 5.5 solved all $10$ of $10$ trials that measured three or more distinct $x_2$ positions across $W$ ($100\%$), compared with $1$ of $5$ with two positions ($20\%$) and $0$ of $17$ with one position ($0\%$; $p = 2 \times 10^{-7}$). Discovery therefore succeeded reliably whenever the hypothesis chain had enough remaining rounds to test and refine the new peak term. When a $\ge 10\sigma$ signal arrived by Round 9 ($11$ trials), Sonnet 5.5 had time to propose a peak term, measure across $W$, and solve $9$ of the $11$ trials ($82\%$). Late hits in Rounds 10--12 left too few rounds for follow-up ($0$ of $8$ solved; $p = 7 \times 10^{-4}$).

Finally, comparing where both agents placed all $768$ measurements across the $32$ paired trials shows how broader coverage helped Sonnet 5.5 reach $W$ while still leaving finer gaps untested (Fig.~\ref{fig:three}c). Both agents chose integer values of $x_2$ in roughly four-fifths of their experiments (Methods): GPT-6.1-Sol concentrated $70\%$ of its measurements at $x_2 = 0$ and $x_2 = 2$, whereas Sonnet 5.5 swept across integers from $0$ to $10$. That broader integer sweep enabled Sonnet 5.5 to enter $19$ of the $23$ peak regions containing an integer ($83\%$), while missing the $9$ narrower regions that fell strictly between consecutive integers ($0/9$; $1/23$ and $0/9$ for GPT-6.1-Sol; Fig.~\ref{fig:three}d).

\section*{Discussion}

Preferring simpler models and testing where models disagree are core principles of quantitative science\cite{chamberlin1890method,platt1964strong,box1967discrimination,schwarz1978estimating}, yet combining them in a closed loop can create a blind spot. Favoring simplicity keeps unsupported terms out of a model, and testing where models differ focuses experiments on informative conditions. When early data miss a narrow feature, however, these two principles work against discovery\cite{sugiyama2006active,sloman2022characterizing,tang2026ridea}: complexity penalties disfavor extra peak parameters, and the leading smooth curves agree across the unmeasured gap. Testing where those curves disagree then pulls experiments toward minor parameter differences of an incomplete background model, much as confirmation and ``streetlight'' biases keep human researchers testing familiar ground\cite{wason1960failure,klayman1987confirmation,kaplan1964conduct}. Asking the right question determines whether a chain of hypotheses ever encounters the evidence needed to revise itself.

Our results also show that current AI scientists already revise equations effectively once experiments expose an anomaly: whenever Sonnet 5.5 measured three or more positions across $W$, it recovered the true law in all $10$ of $10$ trials. The bottleneck lay in starting and sustaining that chain of hypotheses. Focusing tightly on equation disagreement kept GPT-6.1-Sol retesting the smooth background, while coarse integer steps reached only integer-aligned peaks and often arrived too late for follow-up. These findings point to a two-stage strategy for stronger autonomous discovery: spreading early experiments across wide unmeasured intervals to catch unexpected features early\cite{johnson1990minimax,mckay1979comparison,santner2003design}, then concentrating follow-up measurements around anomalies so new hypotheses can be refined. Discovery benchmarks\cite{kabra2026autoscilab,gandhi2025boxinggym,zheng2025newtonbench,duan2025scigym} can likewise accelerate progress by evaluating where agents choose to measure alongside the equations they propose, using non-integer feature locations and simple space-filling baselines.

Although we focused here on a controlled three-variable system without code execution to isolate how hypotheses guide search, the same principle applies broadly across experimental science. Pairing strong language-model hypothesis generators with deliberate gap-probing experimentation offers a direct path toward AI scientists capable of uncovering unexpected phenomena in higher-dimensional and laboratory settings.

\section*{Methods}

\subsection*{Task design and trial setup}

To test whether AI scientists can find a narrow feature inside an unmeasured region, we asked them to study a synthetic system $y = f(x_1, x_2, x_3) + \varepsilon$ with Gaussian measurement noise $\varepsilon \sim \mathcal{N}(0, \sigma^2)$ of standard deviation $\sigma = 0.08$. In each experiment, the agent could choose values for three continuous inputs: $x_1 \in [0, 50]$, $x_2 \in [0, 10]$, and $x_3 \in [20, 45]$, where $x_3$ had no effect on $y$. Unknown to the agent, $y$ rises and levels off with $x_1$, decays smoothly with $x_2$, and contains a narrow Gaussian peak along $x_2$:
\begin{equation}
  y \;=\; \frac{V x_1}{K + x_1}\left(\frac{1}{1 + x_2/J} + A\exp\!\left[-\frac{(x_2 - c)^2}{2S^2}\right]\right) + \varepsilon,
  \label{eq:interior}
\end{equation}
with $V = 10$, $K = 5$, $J = 2$, $A = 0.279$, and $S \approx 0.149$. At $x_1 = 40$, the peak raises $y$ by up to $31\sigma$ inside the peak region $W = (c - 0.35, c + 0.35]$, which covers $7\%$ of the $x_2$ range, while adding less than $2\sigma$ anywhere outside $W$ at every $x_1$. Without the peak, even the exact smooth curve scores $E = 3.9$--$4.8$ depending on $c$. Meeting the solve criterion ($E \le 1.05$) therefore requires finding the peak.

We constructed 32 paired trials with different peak locations across the interior of $x_2$, and tested both agents on every trial. Peak centers $c$ were drawn one from each of 32 equal-width intervals across $[2.5, 8.5]$ in a fixed random order ($23$ peak regions contained an integer, and $9$ lay strictly between consecutive integers). Every trial started with four random conditions, one in each quarter of every input range, redrawn until none fell in $W$ (starting peak signal $< 1\sigma$), and each condition was measured twice. On each paired trial, Claude Sonnet 5.5 and GPT-6.1-Sol received identical starting data and shared the same random seed for measurement noise. We also ran $32$ additional Sonnet 5.5 trials with new peak locations and starting conditions generated by the same stratified procedure, giving $64$ Sonnet 5.5 trials in total ($32$ of which entered $W$) for the post-entry recovery analysis (Fig.~\ref{fig:three}a,b).

\subsection*{Closed-loop experimental protocol}

Both agents received the same short instructions stating only the three input ranges, the noise level $\sigma = 0.08$, and the goal of finding the equation for $y$. Each trial ran for 12 rounds of hypothesis and measurement. At the start of each round, the agent received a text summary of all conditions measured so far (input values, mean $y$, and replicate count) together with up to 12 best-fitting candidate equations from earlier rounds (none in Round 1). Candidate equations were ranked from lowest to highest Bayesian information criterion ($\mathrm{BIC} = \chi^2 + k\ln N$, where $\chi^2$ is the noise-weighted sum of squared errors across $N$ measurements and $k$ is the number of adjustable parameters). Lower BIC favors accurate fits while penalizing extra parameters. For each equation, the agent saw fitted parameter values, prediction errors at each measured condition in units of $\sigma$, and the Python formula.

In each round, the agent briefly explained its reasoning, proposed up to three new equations, and selected two new input conditions to measure next, each measured twice. Every proposed equation had to include between $1$ and $10$ adjustable parameters, together with a starting guess and allowed range for each parameter. Invalid formulas or formulas outside this $1$--$10$ parameter range were skipped before fitting. Across all trials, this occurred on $126/721$ proposals from GPT-6.1-Sol ($17\%$, all zero-parameter formulas with fixed constants) and $5/2{,}495$ from Sonnet 5.5 ($0.2\%$); GPT-6.1-Sol also proposed no new equations in $52$ rounds. At the end of each round, all equations fitted so far had their parameters refitted to the full dataset by bounded nonlinear least squares. Fits started from the agent's parameter guess on first fit or the previous best parameters on refits, plus three random restarts within the allowed ranges. After Round 12, one final step allowed up to three final equations without new measurements, and the lowest-BIC equation was scored across $4{,}096$ fixed uniform random test points ($E = \mathrm{RMSE}/\sigma \le 1.05$).

\subsection*{AI agent setup}

Claude Sonnet 5.5 (Anthropic) and GPT-6.1-Sol (OpenAI) were run at high reasoning effort. Each round was sent as an independent prompt containing the accumulated measurements and ranked candidate equations, and neither agent ran code, used a calculator, or called external tools. If a reply could not be parsed, the agent was prompted once more (only one retry occurred across all $1{,}248$ steps), and every round produced valid conditions.

\subsection*{Search baselines, equation-disagreement metrics, and statistics}

Across the $32$ paired trials, to evaluate where agents searched regardless of what equations they wrote, we recorded whether at least one chosen condition fell in $x_2 \in W$ by round $k$ (for any $x_1$ and $x_3$). We also recorded entry into the central half of $W$ ($|x_2 - c| \le 0.175$; $11/32$ trials for Sonnet 5.5 versus $0/32$ for GPT-6.1-Sol, exact McNemar $p = 0.001$). Across all $768$ measured $x_2$ values per agent, $78\%$ (Sonnet 5.5) and $84\%$ (GPT-6.1-Sol) were integers. We compared entry into $W$ against two search strategies that fit no equations. Random sampling draws $x_2$ uniformly from $[0, 10]$ and enters a $7\%$ region by round $k$ with probability $1 - (1 - 0.07)^{2k}$ ($58\%$ by Round 6 and $82\%$ by Round 12). Starting from each trial's four initial $x_2$ values and the endpoints $0$ and $10$, interval bisection places two measurements per round sequentially at the midpoint of the widest unmeasured $x_2$ interval ($30/32$ entered by Round 6 and $32/32$ by Round 12).

To see how the agents' own equations guided search before the peak was encountered, we analyzed all rounds before the first measurement in $W$ across the $32$ paired trials. First, the top-ranked equation predicted a local peak of at least $2\sigma$ along $x_2$ (at $x_1 = 40$) in only $2/283$ ($0.7\%$) such rounds for Sonnet 5.5 and $2/380$ ($0.5\%$) for GPT-6.1-Sol ($0$ inside $W$ for both). Proposed formulas subtracting a center parameter from $x_2$ (such as $x_2 - c$) appeared $5$ and $7$ times before entering $W$, versus $144$ times for Sonnet 5.5 after entering $W$. Second, in rounds before entering $W$ with at least two distinct fitted equations ($n = 251$ rounds across $31$ trials for Sonnet 5.5; $n = 348$ across $32$ trials for GPT-6.1-Sol), we measured where the top three equations disagreed most. On a reference grid of $201$ evenly spaced $x_2$ values crossed with $x_1 \in \{40, 50, 25, 10, 2\}$ and $x_3 \in \{32.5, 20, 45\}$, we computed the prediction spread (maximum minus minimum, using fitted parameters) among the top three equations. Dividing the larger spread at the agent's two chosen conditions by the grid maximum (capped at $1$) gave round-level medians of $0.89$ for GPT-6.1-Sol and $0.49$ for Sonnet 5.5. Across all $32$ Sonnet 5.5 trials that entered $W$ (Fig.~\ref{fig:three}a,b), each measurement's peak signal was the noiseless peak height above the smooth background at the chosen $(x_1, x_2)$, in units of $\sigma$. A peak term counted as entering the top-ranked equation once the lowest-BIC equation predicted a local peak of prominence $\ge 5\sigma$ inside $W$ along $x_2$ at $x_1 = 40$. By this rule, a peak term entered in $16/19$ trials with a $\ge 10\sigma$ signal ($15$ during Rounds 1--12 and $1$ in the final step), and in $0/11$ unsolved trials whose initial signals stayed below $10\sigma$.

Trial-level percentages are shown with $95\%$ Wilson confidence intervals, and round-level percentages (Fig.~\ref{fig:two}c) with $95\%$ bootstrap confidence intervals from resampling whole trials $10{,}000$ times. Entry rates were compared against random sampling using one-sided exact binomial tests, binary outcomes between agents across the $32$ paired trials using two-sided exact McNemar tests, categorical recovery rates across entered trials using two-sided Fisher exact tests, and continuous per-trial scores using two-sided Wilcoxon signed-rank tests.

\section*{Acknowledgments}

This work was supported by the National Institutes of Health (R01HG014004).

\section*{Competing interests}

The author declares no competing interests.

{\linespread{1.00}
}


\small
\begin{thebibliography}{99}
\raggedright
\setlength{\itemsep}{0pt}
\setlength{\parsep}{0pt}
\setlength{\parskip}{0pt}

\bibitem{chamberlin1890method}
Chamberlin, T. C. The method of multiple working hypotheses. \textit{Science} \textbf{15}, 92--96 (1890).

\bibitem{platt1964strong}
Platt, J. R. Strong inference. \textit{Science} \textbf{146}, 347--353 (1964).

\bibitem{king2009automation}
King, R. D. et al. The automation of science. \textit{Science} \textbf{324}, 85--89 (2009).

\bibitem{wang2023scientific}
Wang, H. et al. Scientific discovery in the age of artificial intelligence. \textit{Nature} \textbf{620}, 47--60 (2023).

\bibitem{boiko2023coscientist}
Boiko, D. A., MacKnight, R., Kline, B. \& Gomes, G. Autonomous chemical research with large language models. \textit{Nature} \textbf{624}, 570--578 (2023).

\bibitem{szymanski2023alab}
Szymanski, N. J. et al. An autonomous laboratory for the accelerated synthesis of novel materials. \textit{Nature} \textbf{624}, 86--91 (2023).

\bibitem{lu2024aiscientist}
Lu, C. et al. The AI Scientist: towards fully automated open-ended scientific discovery. Preprint at \url{https://arxiv.org/abs/2408.06292} (2024).

\bibitem{gottweis2025coscientist}
Gottweis, J. et al. Towards an AI co-scientist. Preprint at \url{https://arxiv.org/abs/2502.18864} (2025).

\bibitem{kabra2026autoscilab}
Kabra, S., Abhyankar, N., Desai, S., Iyer, P. \& Reddy, C. K. LLM-AutoSciLab: closed-loop scientific discovery via active experimentation with LLMs. Preprint at \url{https://arxiv.org/abs/2605.24043} (2026).

\bibitem{schmidt2009distilling}
Schmidt, M. \& Lipson, H. Distilling free-form natural laws from experimental data. \textit{Science} \textbf{324}, 81--85 (2009).

\bibitem{cranmer2023interpretable}
Cranmer, M. Interpretable machine learning for science with PySR and SymbolicRegression.jl. Preprint at \mbox{\url{https://arxiv.org/abs/2305.01582}} (2023).

\bibitem{romera2024funsearch}
Romera-Paredes, B. et al. Mathematical discoveries from program search with large language models. \textit{Nature} \textbf{625}, 468--475 (2024).

\bibitem{novikov2025alphaevolve}
Novikov, A. et al. AlphaEvolve: a coding agent for scientific and algorithmic discovery. Preprint at \mbox{\url{https://arxiv.org/abs/2506.13131}} (2025).

\bibitem{lindley1956measure}
Lindley, D. V. On a measure of the information provided by an experiment. \textit{Ann. Math. Stat.} \textbf{27}, 986--1005 (1956).

\bibitem{box1967discrimination}
Box, G. E. P. \& Hill, W. J. Discrimination among mechanistic models. \textit{Technometrics} \textbf{9}, 57--71 (1967).

\bibitem{atkinson1975design}
Atkinson, A. C. \& Fedorov, V. V. The design of experiments for discriminating between two rival models. \textit{Biometrika} \textbf{62}, 57--70 (1975).

\bibitem{seung1992query}
Seung, H. S., Opper, M. \& Sompolinsky, H. Query by committee. In \textit{Proc. Fifth Annual Workshop on Computational Learning Theory} 287--294 (ACM, 1992).

\bibitem{chaloner1995bayesian}
Chaloner, K. \& Verdinelli, I. Bayesian experimental design: a review. \textit{Stat. Sci.} \textbf{10}, 273--304 (1995).

\bibitem{settles2009active}
Settles, B. \textit{Active Learning Literature Survey}. Computer Sciences Technical Report 1648 (Univ. Wisconsin--Madison, 2009).

\bibitem{rainforth2024modern}
Rainforth, T., Foster, A., Ivanova, D. R. \& Bickford Smith, F. Modern Bayesian experimental design. \textit{Stat. Sci.} \textbf{39}, 100--114 (2024).

\bibitem{sugiyama2006active}
Sugiyama, M. Active learning in approximately linear regression based on conditional expectation of generalization error. \textit{J. Mach. Learn. Res.} \textbf{7}, 141--166 (2006).

\bibitem{sloman2022characterizing}
Sloman, S. J., Oppenheimer, D. M., Broomell, S. B. \& Shalizi, C. R. Characterizing the robustness of Bayesian adaptive experimental designs to active learning bias. Preprint at \url{https://arxiv.org/abs/2205.13698} (2022).

\bibitem{tang2026ridea}
Tang, R., Sloman, S. J. \& Kaski, S. Representative, informative, and de-amplifying: requirements for robust Bayesian active learning under model misspecification. In \textit{Proc. 29th International Conference on Artificial Intelligence and Statistics (AISTATS)}, \textit{PMLR} \textbf{300}, 3016--3024 (2026).

\bibitem{schwarz1978estimating}
Schwarz, G. Estimating the dimension of a model. \textit{Ann. Stat.} \textbf{6}, 461--464 (1978).

\bibitem{wason1960failure}
Wason, P. C. On the failure to eliminate hypotheses in a conceptual task. \textit{Q. J. Exp. Psychol.} \textbf{12}, 129--140 (1960).

\bibitem{klayman1987confirmation}
Klayman, J. \& Ha, Y.-W. Confirmation, disconfirmation, and information in hypothesis testing. \textit{Psychol. Rev.} \textbf{94}, 211--228 (1987).

\bibitem{kaplan1964conduct}
Kaplan, A. \textit{The Conduct of Inquiry: Methodology for Behavioral Science} (Chandler, 1964).

\bibitem{johnson1990minimax}
Johnson, M. E., Moore, L. M. \& Ylvisaker, D. Minimax and maximin distance designs. \textit{J. Stat. Plan. Inference} \textbf{26}, 131--148 (1990).

\bibitem{mckay1979comparison}
McKay, M. D., Beckman, R. J. \& Conover, W. J. A comparison of three methods for selecting values of input variables in the analysis of output from a computer code. \textit{Technometrics} \textbf{21}, 239--245 (1979).

\bibitem{santner2003design}
Santner, T. J., Williams, B. J. \& Notz, W. I. \textit{The Design and Analysis of Computer Experiments} (Springer, 2003).

\bibitem{gandhi2025boxinggym}
Gandhi, K. et al. BoxingGym: benchmarking progress in automated experimental design and model discovery. Preprint at \url{https://arxiv.org/abs/2501.01540} (2025).

\bibitem{zheng2025newtonbench}
Zheng, T. et al. NewtonBench: benchmarking generalizable scientific law discovery in LLM agents. Preprint at \url{https://arxiv.org/abs/2510.07172} (2025).

\bibitem{duan2025scigym}
Duan, H. et al. Measuring scientific capabilities of language models with a systems biology dry lab. In \textit{Adv. Neural Inf. Process. Syst.} \textbf{38} (2025).

\end{thebibliography}
\end{document}